\pdfoutput=1
\documentclass[11pt,a4paper]{article}
\usepackage[hyperref]{eacl2021}
\usepackage{times}
\usepackage{latexsym}

\usepackage{caption}
\usepackage{booktabs}
\usepackage{amsmath,amsfonts,amssymb}
\usepackage{algorithm}
\usepackage[noend]{algorithmic}
\usepackage{xspace}
\usepackage{multirow}
\usepackage{enumitem}
\usepackage{bm}
\usepackage{array}
\usepackage{float}
\usepackage{pgfplots}
\usepackage{tikz}
\usepackage{forest}
\usepackage{adjustbox}
\usepackage{tcolorbox}
\usepackage{textcomp}
\usepackage{comment}
\usepackage{nccmath} 

\usetikzlibrary{arrows,decorations.markings,patterns}
\usetikzlibrary{arrows.meta}
\forestset{edges/.style={for tree={edge={<-,>=stealth}}}}
\usepgfplotslibrary{colorbrewer}    
\pgfplotsset{cycle list/RdBu-4}
\pgfplotsset{compat=1.16}

\def\argmax{\operatornamewithlimits{arg\,max}}

\definecolor{one}{HTML}{222222}
\definecolor{two}{HTML}{56c1ff}
\definecolor{three}{HTML}{ff968d}
\definecolor{four}{HTML}{bdcc2f}
\definecolor{five}{HTML}{9B9B9B}
\definecolor{six}{HTML}{FB8B24}
\definecolor{seven}{HTML}{D81159}

\usepackage{microtype}

\aclfinalcopy 

\newcommand{\fourforums}{\textsc{4Forums}\xspace}

\newcommand{\extref}[2]{\hyperref[{#1}]{\ref*{#1}#2}}

\title{Randomized Deep Structured Prediction for Discourse-Level Processing}

\author{
  Manuel Widmoser\thanks{~~Contributed equally to this work as first authors}~~$^1$\,\,
  Maria Leonor Pacheco\footnotemark[1]~~$^2$ \,\,
  Jean Honorio$^2$ \,\, 
  Dan Goldwasser$^2$ \\[0.25em]
  $^1$University of Salzburg, Austria \\
  $^2$Purdue University, USA \\
  \texttt{mwidmoser@cs.uni-salzburg.at}, \\ \texttt{\{pachecog, jhonorio, dgoldwas\}@purdue.edu} 
}

\date{}

\begin{document}
\maketitle
\begin{abstract}
  Expressive text encoders such as RNNs and Transformer Networks have been at the center of NLP models in recent work. Most of the effort has focused on sentence-level tasks, capturing the dependencies between words in a single sentence, or pairs of sentences. However, certain tasks, such as argumentation mining, require accounting for longer texts and complicated structural dependencies between them. Deep structured prediction is a general framework to combine the complementary strengths of expressive neural encoders and structured inference for highly structured domains. Nevertheless, when the need arises to go beyond sentences, most work relies on combining the output scores of independently trained classifiers. One of the main reasons for this is that constrained inference comes at a high computational cost. In this paper, we explore the use of randomized inference to alleviate this concern and show that we can efficiently leverage deep structured prediction and expressive neural encoders for a set of  tasks involving complicated argumentative structures. 
\end{abstract}

\section{Introduction}\label{sec:intro}
 
Many discourse-level NLP tasks require modeling complex interactions between multiple sentences, paragraphs or even documents. For example, analyzing opinions in online conversations~\cite{hasan2013,sridhar:acl15} requires modeling the dependencies between the opinions in individual posts, the disagreement between posts in long conversational threads and the overall view of users, given all their posts.  

Learning in these settings is extremely challenging. It requires highly expressive models that can capture the claims made in each document, either by using a rich, manually crafted feature set, or by using neural architectures to learn an expressive representation~\cite{ji2014representation,niculae-etal-2017-argument}. In addition, reasoning about the interaction between these decisions is often computationally challenging, as it requires incorporating domain-specific constraints into the search procedure, making exact inference intractable. As a result, most current work relies on highly engineered solutions, which are difficult to adapt. Instead of training structured predictors that model the interaction between decisions during training, they combine locally trained classifiers at test time~\cite{Stab2017PAS}.
 
Our goal in this paper is to study realistic settings, in which discourse-level problems can be learned efficiently when leveraging \textit{deep structured prediction}, a framework for combining rich neural representation with an inference-layer, forcing consistency between them~\cite{zheng2015conditional}.  These models were applied successfully to simpler NLP tagging tasks~\cite{lample2016neural}, in which inference is tractable. However, as shown in a recent argumentation mining work~\cite{niculae-etal-2017-argument}, their applicability to more complex learning tasks is not guaranteed. 

Randomized inference algorithms have been proposed for structured NLP tasks, such as tagging and dependency parsing, in the context of linear models~\cite{zhang2014,zhang-etal-2015-randomized,ma-randomized-19}. This approach offers an efficient alternative to exact inference. Instead of finding the optimal output state, the algorithm makes greedy updates to a randomly initialized (or locally initialized) output assignment state. Our main contribution is to explore these ideas in the context of deep structured models composed of expressive text encoders, where theoretical guarantees are weak or nonexistent. Moreover, we do this for discourse-level tasks involving a rich set of domain constraints. To do this, we consider two variations of this approach. In the first, the algorithm samples and traverses only legal states (i.e., consistent with the constraints imposed by domain knowledge). In the second, these restrictions are ignored and only applied at test time. Adapting the sampling procedure to the specific constraints imposed by each domain is difficult, motivating the second approach as a generic alternative.
    
We focus on two discourse-level tasks, stance prediction in discussion forums, described above, and parsing argumentation structures in essays~\cite{Stab2017PAS}. The latter consists of constructing an argumentation tree that represents the type-of, and relation-between, the arguments made in the essay. Models for both tasks typically employ declarative inference for incorporating domain knowledge.
Our experiments are designed to quantify the trade-off between different modeling choices, both in terms of task performance and computational cost. We compare exact ILP models, approximate inference based on the popular  AD$^3$ algorithm~\cite{martins2015-ad3} and the two randomized inference algorithms. Our experiments show that in all cases, deep structured prediction outperforms traditional shallow approaches, structured learning outperforms inference over locally trained models, and generic randomized inference performs competitively to exact inference.

\section{Related Work}\label{sec:related}


Using deep structured prediction for NLP has been studied in previous work, typically on traditional sentence-level tasks such as dependency parsing~\cite{chen2014fast,weiss-EtAl:2015:ACL-IJCNLP}, transition systems~\cite{Andor2016}, named entity recognition~\cite{lample2016neural} and sequence labeling systems~\cite{ma2016end}. In most of these cases, inference is tractable. More recently, some efforts have started to look at incorporating deep structured prediction to discourse tasks such as argument mining~\cite{niculae-etal-2017-argument},  event and temporal relation extraction~\cite{han-etal-2019-joint} and discourse representation parsing~\cite{liu-etal-2019-discourse}. In all of these cases, constrained inference is formulated as an integer linear program and solved either using off-the-shelf optimizers or approximation algorithms like AD$^3$~\cite{martins2015-ad3}. 

Randomized approximation has been introduced as an alternative to exact inference. \citeauthor{zhang2014}~\shortcite{zhang2014} suggest a simple randomized greedy inference algorithm and empirically demonstrate its effectiveness for dependency parsing and other traditional NLP tasks~\cite{zhang-etal-2015-randomized}.
The theoretical results in~\cite{HonorioJ16}, based on the probably approximately correct Bayes framework, characterize these findings by providing generalization bounds.
More recently, \citeauthor{ma-randomized-19}~\shortcite{ma-randomized-19} extended the work of~\cite{zhang2014,zhang-etal-2015-randomized} to structured prediction tasks with large structured outputs by leveraging local classifiers to find good starting solutions and improve the accuracy of search. All of these methods were evaluated on linear structured models.  

In this paper, we focus on two tasks: mining argumentative structures in essays and stance prediction in online debates. \citeauthor{Stab2017PAS}~\shortcite{Stab2017PAS} approach argumentative essays using an exhaustive set of hand-crafted features, linear local classifiers and ILP at test time. \citeauthor{niculae-etal-2017-argument}~\shortcite{niculae-etal-2017-argument} jointly learn to score multiple decisions while enforcing domain constraints. They explore structured SVMs and RNNs, using the AD$^3$ inference algorithm~\cite{martins2015-ad3}. On the other hand, there are several works on predicting user stances in online debates. Some approaches model the problem as a text classification task~\cite{Somasundaran:2010,C18-1203}, while other approaches take a collective approach to model user behavior and interactions~\cite{Walker:2012:SCU:2382029.2382124,hasan2013,sridhar:acl15,li2018}. In the latter case, inference procedures include MaxCut, ILP and probabilistic soft logic~\cite{psl}.






\section{Modeling}\label{sec:modeling}

We look at two challenging structured prediction problems that deal with long texts where dependencies span across different paragraphs, documents and authors. To deal with these setups, we define neural factor graphs $G = \{\Psi\}$ where each decision $\psi_i \in \Psi$ is associated with a neural architecture $\rho_i$ and a set of parameters $\bm\theta_i$. In this section, we introduce the tasks in detail. 


\subsection{Argument Mining}

This task aims to identify argumentative structures in essays. Each argumentative structure forms a tree, and there is a forest per document. Nodes correspond to spans of text in the document and they can be labeled either as \textit{claims, major claims} or \textit{premises}. Edges correspond to stances (i.e., support/attack relations between nodes). The spans of texts are given, and we need to label nodes, predict which pairs of nodes are connected by an edge and label the edges. Domain knowledge can be exploited as there are only valid edges between pairs of premises, a premise and a claim, or a claim and a major claim. At the same time, all trees are rooted at major claims. Similarly to previous work, we model second order relationships: \textbf{grandparent} ($a\rightarrow b\rightarrow c$) and \textbf{co-parent} ($a\rightarrow b\leftarrow c$)~\cite{martins2014,niculae-etal-2017-argument}.

Figure~\ref{fig:modeling} has a visual representation of the structure. In this problem, each forest defines a factor graph $\Psi$ and $G$ is the collection of all documents. We define a set of five neural architectures corresponding to the five types of decisions that we need to make: $NN = \{\rho_{\text{node}}, \rho_{\text{link}}, \rho_{\text{stance}}, \rho_{\text{grandparent}}, \rho_{\text{coparent}}\}$, each with its own set of parameters $\bm\theta = \{ \theta_{\text{node}}, \theta_{\text{link}}, \theta_{\text{stance}}, \theta_{\text{grandparent}}, \theta_{\text{coparent}} \}$. Note that in principle, we can substitute each $(\rho_i, \theta_i)$ with any neural architecture. We include details about the architectures in the experimental section. 

\subsection{Stance Prediction}

\begin{figure}[t]
    \centering
    \begin{minipage}{.485\columnwidth}
        \centering
        \resizebox{\textwidth}{!}{%
            \begin{tikzpicture}
    \definecolor{coparent}{HTML}{56c1ff}
    \definecolor{stance}{HTML}{D81159}
    \definecolor{grandparent}{HTML}{FB8B24}

    \begin{scope}[every node/.style={circle, thick,draw, minimum size=2em}]
        \node (A) at (0,0) {\scriptsize MC};
        \node (B) at (-1,-2) {C};
        \node (C) at (1,-2) {C};
        \node (D) at (-2,-4) {P};
        \node (E) at (0,-4) {P};
        \node (F) at (2,-4) {P} ;

        \node[rectangle, minimum size=0.5em, inner sep=0pt, coparent, fill=coparent] (G) at (-1,-3.5) {};
        \node[rectangle, minimum size=0.5em, inner sep=0pt, coparent, fill=coparent] (H) at (0,-1.5) {};
    \end{scope}

    \begin{scope}[>={Stealth}]
        \path [->, thick] (B) edge node[fill=stance, rectangle, minimum size=0.5em, inner sep=0pt, anchor=center, pos=0.4] {} (A);
        \path [->, thick] (C) edge node[fill=stance, rectangle, minimum size=0.5em, inner sep=0pt, anchor=center, pos=0.4] {} (A);
        \path [->, thick] (D) edge node[fill=stance, rectangle, minimum size=0.5em, inner sep=0pt, anchor=center, pos=0.4] {} (B);
        \path [->, thick] (E) edge node[fill=stance, rectangle, minimum size=0.5em, inner sep=0pt, anchor=center, pos=0.4] {} (B);
        \path [->, thick] (F) edge node[fill=stance, rectangle, minimum size=0.5em, inner sep=0pt, anchor=center, pos=0.4] {} (C);

        \path [->, thick, draw=grandparent, grandparent] (A) edge[bend right=30]  node {} (D);
        \path [->, thick, draw=grandparent, grandparent] (A) edge[bend left=30]  node {} (F);
        \path [->, thick, draw=grandparent, grandparent] (A) edge[bend left=15]  node {} (E);

        \path [-, thick] (B) edge[coparent, fill=coparent] node {} (G);
        \path [->, thick] (G) edge[coparent, fill=coparent] node {} (D);
        \path [->, thick] (G) edge[coparent, fill=coparent] node {} (E);

        \path [-, thick] (A) edge[coparent, fill=coparent] node {} (H);
        \path [->, thick] (H) edge[coparent, fill=coparent] node {} (B);
        \path [->, thick] (H) edge[coparent, fill=coparent] node {} (C);
    \end{scope}

    \begin{scope}
        \matrix [] at (-1.6,-0.3) {
            \node [rectangle, minimum size=0.5em, inner sep=0pt, fill=black, label=right:{\footnotesize link}] {}; \\[-0.3em]
            \node [rectangle, minimum size=0.5em, inner sep=0pt, fill=stance, label=right:{\footnotesize stance}] {}; \\[-0.3em]
            \node [rectangle, minimum size=0.5em, inner sep=0pt, fill=grandparent, label=right:{\footnotesize grandparent}] {}; \\[-0.3em]
            \node [rectangle, minimum size=0.5em, inner sep=0pt, fill=coparent, label=right:{\footnotesize co-parent}] {}; \\[-0.3em]
        };
    \end{scope}
\end{tikzpicture}
        }\\[0.5em]
    \end{minipage}\hfill 
    \begin{minipage}{.485\columnwidth}
        \centering
        \resizebox{\textwidth}{!}{%
            \begin{tikzpicture}
    \definecolor{agreement}{HTML}{56c1ff}
    \definecolor{con}{HTML}{D81159}
    \definecolor{disagreement}{HTML}{ffffff}

    \begin{scope}[every node/.style={circle, thick,draw, minimum size=2em}]
        \node[pattern=north east lines, pattern color=black] (A) at (0,0) {};
        \node[pattern=north east lines, pattern color=con] (B) at (-1,-2) {};
        \node[pattern=north east lines, pattern color=black] (C) at (1,-2) {};
        \node[pattern=north east lines, pattern color=black] (D) at (-2,-4) {};
        \node[pattern=north east lines, pattern color=con] (E) at (0,-4) {};
        \node[pattern=north east lines, pattern color=black] (F) at (2,-4) {} ;
    \end{scope}

    \begin{scope}[>={Stealth}]
        \path [->, thick] (B) edge node[fill=disagreement, draw=black, rectangle, minimum size=0.5em, inner sep=0pt, anchor=center, pos=0.4] {} (A);
        \path [->, thick] (C) edge node[fill=agreement, draw=black, rectangle, minimum size=0.5em, inner sep=0pt, anchor=center, pos=0.4] {} (A);
        \path [->, thick] (D) edge node[fill=disagreement, draw=black, rectangle, minimum size=0.5em, inner sep=0pt, anchor=center, pos=0.4] {} (B);
        \path [->, thick] (E) edge node[fill=agreement, draw=black, rectangle, minimum size=0.5em, inner sep=0pt, anchor=center, pos=0.4] {} (B);
        \path [->, thick] (F) edge node[fill=agreement, draw=black, rectangle, minimum size=0.5em, inner sep=0pt, anchor=center, pos=0.4] {} (C);
    \end{scope}

    \begin{scope}
        \matrix [] at (1.75,-0.3) {
            \node [rectangle, minimum size=0.5em, inner sep=0pt, fill=black, label=right:{\footnotesize pro}] {}; \\[-0.3em]
            \node [rectangle, minimum size=0.5em, inner sep=0pt, fill=con, label=right:{\footnotesize con}] {}; \\[-0.2em]
            \node [rectangle, minimum size=0.45em, inner sep=0pt, fill=agreement, draw=black, thick, label=right:{\footnotesize agree}] {}; \\[-0.35em]
            \node [rectangle, minimum size=0.45em, inner sep=0pt, fill=disagreement, draw=black, thick, label=right:{\footnotesize disagree}] {}; \\[-0.3em]
        };
    \end{scope}
\end{tikzpicture}
        }\\[0.5em]
    \end{minipage}

    \caption{Argument Mining (left), Stance Prediction (right)}
    \label{fig:modeling}
\end{figure}
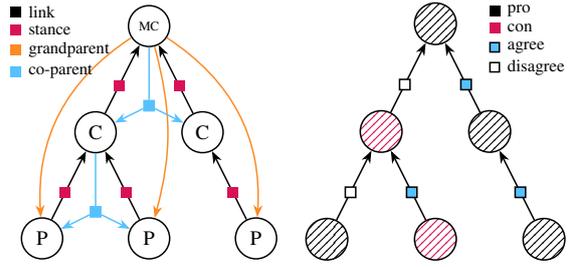

Given a debate thread on a specific political issue, the task is to predict the stance of each post w.r.t. the issue (e.g., pro-life or pro-choice on abortion) ~\cite{Walker:2012:SCU:2382029.2382124}. Following previous work, we model the problem as a collective classification task and consider all posts in a given thread. To do this, we add the task of predicting stance agreement between consecutive posts. As observed in Figure~\ref{fig:modeling}, each thread forms a tree, where users participate and respond to each other's posts. For a thread labeling to be valid, we need to enforce consistency between the node and edge labels. 

In this case, each discussion thread defines a factor graph $\Psi$ and $G$ is the collection of threads. We define two neural architectures $NN = \{\rho_{\text{stance}}, \rho_{\text{agreement}}\}$, each with its own set of parameters $\bm\theta = \{ \theta_{\text{stance}}, \theta_{\text{agreement}}\}$. As in the previous setup, each $(\rho_i, \theta_i)$ can be substituted by any neural architecture, more details are outlined in the experimental section. 


\section{Learning}\label{sec:learning}

We learn a joint neural model that uses inference during training to ensure consistency across all decisions. Let $\Psi$ be a factor graph with potentials $\psi_i \in  \Psi$ over all possible structures $Y$. Let $\bm{x_i}$ be the input vector to potential $\psi_i$. Let  $\bm{\theta} = \{ \theta^i\}$ be a set of parameter vectors associated with a set of neural networks $\bm{\rho} = \{ \rho_{i}\}$, and $\rho_{i}(\bm{x_i,y_i}; \theta^i)$ is the score for potential $\psi_i$ resulting from a forward pass. Here $\bm{y} \in Y$ corresponds to the gold structure and $\bm{\hat{y}} \in Y$ to the prediction resulting from the MAP inference procedure:


\begin{equation}
\begin{split}
\argmax_{\bm{y}\in Y} \sum_{\psi_i \in \Psi} \rho_{i}(\bm{x_i,y_i}; \theta^i) \\
s.t. \; c(\bm{x_c,y_c}) \;\;\forall c \in C
\end{split}
\label{eq:inference}
\end{equation}


\noindent Where $C$ is a set of domain-specific constraints defined over the factor graph $\Psi$, and $\bm{x_c}, \bm{y_c}$ indicates inputs and variables relevant to the constraints. In this work, we experiment with different algorithms to obtain or approximate the $\argmax$, including the randomized procedures outlined in Section~\ref{sec:inference}. 

To learn $\bm{\theta}$, we use the structured hinge loss $L(\bm{x,y,\hat{y};\theta})$ defined as: 

\begin{equation}\fontsize{10}{12}\selectfont
\begin{split}
\max\Bigl(0, \, \max_{\hat{\textbf{y}} \in Y}\bigl(\Delta(\bm{y},\hat{\bm{y}}) + \sum_{\psi_i \in \Psi} \rho_i(\bm{x_i,\hat{y}_i}; \theta^i)\bigr)   \\ - \sum_{\psi_i \in \Psi} \rho_i(\bm{x_i,y_i}; \theta^i)\Bigr)
\end{split}
\label{eq:structured_hinge_loss}
\end{equation}


\noindent Where $\Delta(\bm{y},\hat{\bm{y}})$ is the Hamming loss. To introduce the Hamming loss into the objective, we perform loss augmented inference. The pseudo-code for the structured learning procedure can be observed in Algorithm~\ref{alg:learning}. We implemented our models using DRaiL \cite{pacheco2020modeling}, a declarative deep structured prediction framework built on PyTorch, and extended it to support our randomized inference procedures\footnote{The source code for this paper is available on \newline \url{https://www.gitlab.com/purdueNlp/DRaiL}}.

\begin{algorithm}[ht]
    {\fontsize{10}{13}\selectfont
    \begin{algorithmic}[1]
     \STATE $p \gets 0$
     \STATE $\text{loss}_\text{best} \gets \infty$
     \STATE $\bm\theta \gets \bm\theta^{\text{local}}$
     \STATE $\bm\theta^{\text{ret}} \gets \bm\theta$

    \WHILE{$p <$ patience}
        \FOR{\textbf{each} $\Psi \in G_{\text{train}}$}
            \FOR{\textbf{each} $\psi_i \in \Psi$}
                \STATE $w_i \gets \rho_i(\bm{x_i,y_i}; \theta^i)$ \;\;\;\;\;\;\;\;\; \textit{// forward pass}
            \ENDFOR
            \STATE $\hat{\bm{y}} \gets \argmax_{\bm{y}\in Y} \sum_{\psi_i \in \Psi} w_{i}\psi_{i}(\bm{x_i,y_i})$
            \STATE $\text{loss} \gets L(\bm{x,y,\hat{y};\theta})$
            \STATE backpropagate loss

        \ENDFOR
        
        \STATE $\text{loss}_\text{dev} \gets 0$
        \FOR{\textbf{each} $\Psi \in G_{\text{dev}}$}
            \FOR{\textbf{each} $\psi_i \in \Psi$}
                \STATE $w_i \gets \rho_i(\bm{x_i,y_i}; \theta^i)$ \;\;\;\;\;\;\;\;\; \textit{// forward pass}
           \ENDFOR
            \STATE $\hat{\bm{y}}_{\text{dev}} \gets \argmax_{\bm{y}\in Y} \sum_{\psi_i \in \Psi} w_{i}\psi_{i}(\bm{x_i,y_i})$
            \STATE $\text{loss}_\text{dev}  \gets \text{loss}_\text{dev} + L(\bm{x,y_\text{dev},\hat{y}_\text{dev};\theta})$
        \ENDFOR
        \IF{$\text{loss}_\text{dev} < \text{loss}_\text{best}$}
            \STATE $\text{loss}_\text{best} \gets \text{loss}_\text{dev}$
            \STATE $\bm\theta^{\text{ret}} \gets \bm\theta$
            \STATE $p \gets 0$
        \ELSE
            \STATE $p \gets p + 1$
    \ENDIF

     \ENDWHILE
     \RETURN $\bm\theta^{\text{ret}}$

    \end{algorithmic}
    }
    \caption{Deep Structured Prediction}
    \label{alg:learning}
    
\end{algorithm}

\section{Randomized Inference}
\label{sec:inference}

In this section, we describe the randomized inference procedure used for each task. We define the relevant domain constraints for each case, and explain how we sample solutions that respect them. Finally, we include a discussion about the theoretical bounds for the linear case. 

\subsection{Argument Mining}

For randomized inference on argument mining, we adapt the randomized greedy algorithm proposed by ~\citeauthor{zhang2014}~\shortcite{zhang2014}. Algorithm~\ref{alg:randinference} outlines the overall procedure. We will consider that each paragraph $p \in P$ of an essay contains a single tree. We obtain a local optimum tree $\hat{y}$ by using the hill climbing algorithm, which is further described below. After that, $\hat{y}$ is labeled and added to the forest $Y$. We iterate over each paragraph (line 4) and subsequently score the forest as:

\begin{equation}
    \mathcal{\bar{S}}(Y) = \sum_{\hat{y} \in Y} \mathcal{S}(\hat{y}) = \sum_{\hat{y} \in Y} w + h \left\Vert y - \hat{y} \right\Vert_1
\end{equation}

\noindent Where $w = \sum_{\psi_i \in \Psi} \rho_i(\bm{x_i,\hat{y}_i};\theta^i)$ is the sum of the scores of the potentials for the predicted tree $\hat{y}$. 
We add a weighted Hamming distance term to the scoring function in order to additionally penalize the score the more the tree structure differs from the gold structure.
$h \left\Vert y - \hat{y} \right\Vert_1$ gets close to $w$ if the distance is low, and close to zero if it is high.
More specifically, let $\left\Vert y - \hat{y} \right\Vert_1$ be in $[0,1]$, e.g., by dividing the number of node and edge differences by the total number of nodes and edges.
In its simplest form, $h$ can be assigned to $-w$, and thus $\mathcal{S}(\hat{y})$ would become $w$ if $y = \hat{y}$ or $0$ if they differ in every node and edge.
Whenever the score of the locally improved forest is better than the forest found so far, $Y$ becomes the new currently best scoring forest $\hat{Y}$.
Since hill climbing might get stuck in a local optimum, we repeat line 3-9 for a constant number of restarts.

\begin{algorithm}[t]
    {\fontsize{10}{12}\selectfont
    \begin{algorithmic}[1]
        \STATE $\hat{Y} \gets \{\}$

        \FOR {\text{number of restarts}}
            \STATE $Y \gets \{\}$

            \FOR {\textbf{each} $p \in P$}
                \STATE $\hat{y} \gets \text{hill\_climbing}(p)$
                \STATE $\text{label}(\hat{y})$
                \STATE $Y \gets Y \cup \{ \hat{y} \}$
            \ENDFOR

            \IF {$\mathcal{\bar{S}}(Y) > \mathcal{\bar{S}}(\hat{Y})$}
                \STATE $\hat{Y} \gets Y$
            \ENDIF
        \ENDFOR
        \RETURN $\hat{Y}$
    \end{algorithmic}
    }
    \caption{Randomized Inference}
    \label{alg:randinference}
\end{algorithm}

\begin{algorithm}[ht]
    {\fontsize{10}{12}\selectfont
    \begin{algorithmic}[1]
        \STATE $\hat{y}_0 \gets \text{initialize tree randomly for paragraph } p$
        \STATE $\text{label}(\hat{y}_0)$
        \STATE $\hat{y} \gets \hat{y}_0$
        \STATE $t \gets 0$

        \REPEAT
            \STATE $\mathcal{L} \gets \text{top-down level node list of $\hat{y}$}$
            \FOR {$i=1,...,|\,\mathcal{L}\,|$}
                \FOR {$j=i-1,...,0$}
                    \STATE $\hat{y}_{t+1} \gets \text{connect subtree of $\mathcal{L}_i$ to $\mathcal{L}_j$}$
                    \STATE $\text{label}(\hat{y}_{t+1})$
                    
                    \IF {$\mathcal{S}(\hat{y}_{t+1}) > \mathcal{S}(\hat{y})$}
                    \STATE $\hat{y} \gets \hat{y}_{t+1}$
                    \ENDIF
                    
                    \STATE $t \gets t + 1$
                \ENDFOR
            \ENDFOR
        \UNTIL {\text{no improvement in this iteration}}
        \RETURN $\hat{y}$
    \end{algorithmic}
    }
    \caption{Hill Climbing}
    \label{alg:hillclimbing}
\end{algorithm}

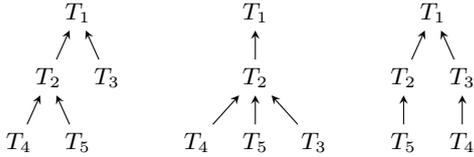
\begin{figure}[t]\small
    \centering
    \begin{adjustbox}{valign=t}
        \begin{forest}
            edges
            [$T_1$
                [$T_2$
                    [$T_4$]
                    [$T_5$]
                ]
                [$T_3$]
            ]
        \end{forest}
    \end{adjustbox}\qquad
    \begin{adjustbox}{valign=t}
        \begin{forest}
            edges
            [$T_1$
                [$T_2$
                    [$T_4$]
                    [$T_5$]
                    [$T_3$]
                ]
            ]
        \end{forest}
    \end{adjustbox}\qquad
    \begin{adjustbox}{valign=t}
        \begin{forest}
            edges
            [$T_1$
                [$T_2$
                    [$T_5$]
                ]
                [$T_3$
                    [$T_4$]
                ]
            ]
        \end{forest}
    \end{adjustbox}
    \caption{Greedy local update of a tree $\hat{y}_t$ (left) to $\hat{y}_{t+1}$ and $\hat{y}_{t+2}$ without score improvement}
    \label{fig:localupdates}
\end{figure}

Algorithm~\ref{alg:hillclimbing} describes the hill climbing procedure. It initially draws uniformly a tree $\hat{y}_0$ at random.
Then the greedy algorithm applies local updates on $\hat{y}_t$ and attempts to achieve a better scoring tree $\hat{y}_{t+1}$.
This is done by iterating through a top-down level node list $\mathcal{L}$ of $\hat{y}_t$.
Denote $i$ as the current position in the list, then the entire subtree of $\mathcal{L}_i$ is connected to the node $\mathcal{L}_{j}$, whereas $j=i-1,i-2,\dots,0$.
If the score of $\hat{y}_{t+1}$ is higher than the score of $\hat{y}_t$, the newly generated tree is kept.
The algorithm continues until the score can no longer be improved and therefore yields a local optimum tree.
Figure~\ref{fig:localupdates} depicts how such local updates are performed, $\mathcal{L}=(T_1, T_2, T_3, T_4, T_5)$.

It might be the case that a paragraph contains more than a single tree, therefore, when a tree is initially drawn at random, we introduce an additional \textit{phantom node} which serves as the new root.
This modification no longer restricts hill climbing on trees only.
Moreover, it allows us having multiple roots and we treat the second layer of the tree like the top layer.

\paragraph{Domain Specific Constraints:} For node labeling, we exploit domain knowledge.
\textit{Major claims} can only occur in the first or last paragraph, and there has to be at least one major claim in each essay.
A root gets labeled as major claim with some fixed probability depending on the paragraph (first or last), holding the condition that there has to exist at least one.
Any other root is labeled as a \textit{claim} in each paragraph.
Nodes having an edge to a major claim are labeled as claims as well. 
All remaining nodes are \textit{premises}. An edge can have either the label \textit{support} or \textit{attack} and we draw all edge labels randomly with a probability of $0.9$ being a support label. The node and edge labels are determined after each iteration since scoring depends on both, links and labels.

In Section~\ref{sec:experiments}, we evaluate our models using randomized inference with and without domain specific constraints. In the latter case, all labels are chosen at random.

\subsection{Stance Prediction}

A debate thread provides a fixed structure, thus nodes and links are predefined and no improvement of the tree structure needs to be done.
However, nodes and edges still need to be labeled and can be improved.
Initially, we pick the node labels, which can either be \textit{pro} or \textit{con}. Following the observations made by \citeauthor{ma-randomized-19}~\shortcite{ma-randomized-19}, we leverage local classifiers and greedily chose the label with the highest score for each node. 

\paragraph{Domain Specific Constraints:} To respect the dependencies between node and edges labels, we use the following heuristic: If two consecutive nodes $u$ and $v$ have different stances, the edge $(u,v)$ receives a \textit{disagreement} label, if they share the same stance, $(u,v)$ gets an \textit{agreement} label. 
When author constraints are considered as well, we additionally force stances of posts to be equal when written by the same author.

We attempt to improve node labels by flipping them randomly and subsequently adjust the edge labels.
This is done until an iteration no longer improves the overall score. We restart the algorithm for a constant number of times in order to increase the chance of achieving a global optimum. In the experiments, we evaluate our models using randomized inference with and without domain specific constraints. When constraints are not used, a random node is flipped and its adjacent edge adjusted, without enforcing consistency in the whole tree. 

The error of the \textit{constrained} randomized algorithms can be bound for the linear case. Let's define the norm of the set of parameter vectors $\bm{\theta}$ as follows: $\|\bm{\theta}\| = \sqrt{\sum_{\theta^i \in \bm{\theta}} \|\theta^i\|^2}$, where $\|\theta^i\|$ is the Euclidean norm of the parameter vector $\theta^i$.
Let $n$ be the number of training samples.
From Theorem 2 and Claim ii in~\cite{HonorioJ16}, for $\rho_{i}(\bm{x_i,y_i}; \theta^i)$ linear in $\theta^i$, the generalization bound (i.e., the difference between the test error and training error) is on the order of $\|\bm{\theta}\|^2/n + \|\bm{\theta}\|/\sqrt{n} + \max(1/\log 2,\|\bm{\theta}\|^2) \log^{3/2} n/\sqrt{n}$.
The above generalization bound is decreasing in $n$, and increasing in $\|\bm{\theta}\|$, which suggests the use of a large training set, and the penalization of the norm $\|\bm{\theta}\|$ during learning. In our experiments, we show that in practice we can obtain competitive results by implementing the randomized algorithms for the non-linear case.
\section{Experiments}\label{sec:experiments}



We learn our models using four different inference procedures: (1) \textbf{ILP} defines the inference problem as an integer linear program and uses the Gurobi solver\footnote{\url{https://www.gurobi.com/products/gurobi-optimizer}} to perform exact inference, (2) \textbf{AD$^3$/ILP} translates the ILP program into an AD$^3$ instance to perform approximate inference, (3) \textbf{Rand-C} uses the randomized method with domain constraints, and (4) \textbf{Rand} uses the randomized method without domain constraints. Note that we always use exact inference to evaluate on both the development and test sets. For completeness, we add an entry \textbf{AD$^3$} where we use AD$^3$ for both training and testing. When using ILP or AD$^3$, the domain constraints are expressed declaratively. 

All experiments were run on a 32 core 3.2Ghz Intel Xeon CPU machine with 128GB RAM and an NVIDIA GeForce GTX 1080 Ti 11GB GDDR5X GPU. We performed an exhaustive search for hyper-parameters on the development set. We tuned the learning rate ($\text{lr}\in$ \{1e-6, 2e-6, 5e-6, 1e-5, 2e-5, 5e-5, 1e-4, 2e-4, 5e-4,  1e-3, 2e-3, 5e-3, 1e-2, 2e-2, 5e-2, 1e-1\}), patience ($\text{p}\in\{5, 10, 15, 20\}$), and number of restarts ($\text{r}\in\{1, 5, 10, 15, 20, 30, 50, 100\}$). The weight decay was fixed at 1e-5 (PyTorch's default). We found that results were stable for local and global models, for different sets of constraints and across inference algorithms.

\subsection{Argument Annotated Persuasive Essays}\label{sec:argmining_exp}

\paragraph{Dataset:} We used the UKP dataset \cite{Stab2017PAS}, consisting of 402 documents,  with  a  total of 6,100  propositions and 3,800 links (17\% of pairs). We use the train/dev/test splits used by \citeauthor{niculae-etal-2017-argument}~\shortcite{niculae-etal-2017-argument}, and report macro F1 for components and positive F1 for relations.

\paragraph{Learning and Representation:} We did 5 repetitions and reported the average performance. Each repetition used a different seed to initialize the model parameters. For training, we used stochastic gradient descent, a patience of 10, weight decay of 1e-5, and 5 restarts for randomized inference. For local models, we used a learning rate of 0.05 and for structured learning we used a learning rate of 1e-4. Similarly to previous work on deep structured prediction \cite{han-etal-2019-joint}, we obtained better results by performing structured learning over locally trained models, instead of training them from scratch. To represent the component and the essay, we used a BiLSTM over the words, initialized with GloVe embeddings \cite{pennington2014glove}, concatenated with a feature vector following  \citeauthor{niculae-etal-2017-argument}~\shortcite{niculae-etal-2017-argument}, without the word features. For representing the relation, we use the components, as well as the relation features used in \citeauthor{niculae-etal-2017-argument}~\shortcite{niculae-etal-2017-argument}. For shallow models, we use a bag-of-words representation for the text and concatenate it with the rest of the features into a single feature vector. Both the feature extraction and the neural implementations are available in the repository. 


We test two versions of the model: (1) \textbf{Base} includes node labeling, link prediction and link labeling, and (2) \textbf{Full} adds grandparent and co-parent factors. Domain constraints are introduced in all models. 


\begin{table}[ht]
    \centering
    \resizebox{\columnwidth}{!}{%
        \begin{tabular}{llcccc}
            \toprule
            \multicolumn{1}{l}{Model} & \multicolumn{1}{l}{Inference} & Node & Link & Avg & Stance  \\
            \midrule
            \multirow{2}{*}{\bf Local} 
             & --  & 70.7 & 52.8 & 61.7 (60.7) & 63.4   \\
             & L+I & 76.5 & 56.9 & 66.7 (66.5) & 62.5  \\
            \midrule
            \multirow{5}{*}{\bf Base} 
                & ILP             & 83.0 & 57.6 & 70.3 (67.2) & 68.0 \\
                & AD$^3$/ILP      & \textbf{83.2} & 58.2 & \textbf{70.7} (67.2) & \textbf{68.4} \\
                & AD$^3$          & 83.0 & 57.6 & 70.3 (67.2) & 68.0 \\
                & Rand-C          & 82.8 & 58.4 & 70.6 (67.7) & \textbf{68.4} \\
                & Rand            & 82.9 & \textbf{58.5} & \textbf{70.7} (67.7) & 68.0 \\
            \midrule
            \multirow{5}{*}{\bf Full} 
                & ILP             & 83.1 & 61.2 & 72.2 (65.3) & \textbf{69.2} \\
                & AD$^3$/ILP      & 83.7 & 62.0 & 72.9 (65.3) & 68.5 \\
                & AD$^3$          & 83.5 & 61.1 & 72.3 (65.3) & \textbf{69.2} \\
                & Rand-C          & \textbf{83.8} & 62.6 & 73.2 (66.3) & 68.4 \\
                & Rand            & 83.4 & \textbf{63.2} & \textbf{73.3} (65.9) & 68.4 \\
            \bottomrule
        \end{tabular}
    }
    \caption{F1 for argument mining using \textbf{deep structured prediction}, Avg results using \textbf{shallow} models included in parenthesis}
    \label{tab:deep_argument}
\end{table}

\noindent We can analyze the results across three dimensions:

\paragraph{Structured Learning:} The advantage of leveraging more structural dependencies can be seen in Table~\ref{tab:deep_argument}. The model gets increasingly better as more dependencies are considered, and using global learning outperforms learning local models and using inference just at prediction time (L+I). 

\paragraph{Deep vs. Shallow:} There is a consistent trend showing that deep structured models are more expressive than their shallow counterparts, as we can see by comparing average results in Table~\ref{tab:deep_argument}. To obtain good results using linear classifiers,  \citeauthor{Stab2017PAS}~\shortcite{Stab2017PAS} relied on an exhaustive set of features (Table~\ref{tab:argument_mining_previous_work}). These numbers cannot be replicated by using just word-features and the feature set suggested by \citeauthor{niculae-etal-2017-argument}~\shortcite{niculae-etal-2017-argument}, as our shallow models and their structured SVM results show. In contrast, deep models and word embeddings are able to leverage this information without additional features. In addition, we find that deep models have a shorter overall training time (3.3x faster for the full model). This can be attributed to the compact embedding representation used in deep models, in contrast to the large sparse one-hot vectors used in linear models. Similarly to previous work \cite{niculae-etal-2017-argument}, we find that higher-order factors and strict constraints are more helpful when using deep structured models than in their shallow counterparts. 

\paragraph{Randomized vs. ILP/AD3:} When using deep structured prediction, we did not find a statistically significant difference in the performance of the models that were trained with ILP/AD3 vs. the ones that were trained with constrained and non-constrained randomized inference.




\begin{table}[t]
    \centering\small
    \resizebox{\columnwidth}{!}{%
        \begin{tabular}{lccc}
            \toprule
            \multicolumn{1}{l}{Model} & Node & Link & Avg \\
            \midrule
            Human upper bound   & 86.8 & 75.5 & 81.2 \\
            \midrule
            ILP Joint \cite{Stab2017PAS} & 82.6 & 58.5 & 70.6 \\ 
            Struct RNN strict  \cite{niculae-etal-2017-argument}  & 79.3 & 50.1 & 64.7 \\
            Struct RNN full \cite{niculae-etal-2017-argument} & 76.9 & 50.4 & 63.6 \\ 
            Struct SVM strict \cite{niculae-etal-2017-argument} & 77.3 & 56.9 & 67.1 \\ 
            Struct SVM full \cite{niculae-etal-2017-argument} & 77.6 & 60.1 & 68.9 \\ 
            Joint PointerNet \cite{potash-etal-2017-heres} & 84.9 & 60.8 & 72.9 \\ 
            \citealt{kuribayashi-etal-2019-empirical} & \textbf{85.7} & \textbf{67.8} & \textbf{76.8} \\
            BERT \cite{devlin2018bert}  & 71.1 & 50.8 & 61.0 \\
            BERT-doc & 79.5 & 55.8 & 67.7 \\
            BERT-doc + Inf (Base) & 79.9 & 58.1 & 69.0 \\ 
            BERT-doc + Structured Prediction (Base) & 82.1 & 60.0 & 71.1 \\ 
            \midrule
            Deep Full ILP    & 83.1 & 61.2 & 72.2 \\
            Deep Full Rand-C & \textbf{83.8} & 62.6 & 73.2 \\
            Deep Full Rand   & 83.4 & \textbf{63.2} & \textbf{73.3} \\

            \bottomrule
        \end{tabular}
    }
    \caption{Previous work on UKP Dataset}
    \label{tab:argument_mining_previous_work}
\end{table}

We obtain competitive results with respect to previous work that relies on the same underlying embeddings or features, as observed in Table~\ref{tab:argument_mining_previous_work}. Recently, \citet{kuribayashi-etal-2019-empirical} were able to further improve performance by exploiting contextualized embeddings that look at the whole document, instead of embedding the arguments in isolation, and by making a distinction between argumentative markers and argumentative components. We attempted document-level contextualized embeddings using BERT and were not able to replicate their success\footnote{We did not experiment with their extended BoW features, nor AC/AM distinction.}. Moreover, we found no significant improvement on the structured prediction models when replacing our BiLSTM encoders with either BERT or document-level BERT. We leave the exploration of an effective way to leverage contextualized embeddings for future work. As for stance prediction,  \citeauthor{Stab2017PAS}~\shortcite{Stab2017PAS} identify stances over the resulting structure and obtain a macro F1 of 68.0. Our full models obtain commensurate results, 69.2, 68.4 for ILP and randomized inference, respectively. 


\subsection{Debate Stance Prediction}\label{sec:4forums_exp}

\paragraph{Dataset:} We use a subset of the \fourforums dataset from the Internet Argument Corpus \cite{Walker:2012:SCU:2382029.2382124}, which consist of a total of 418 discussion threads on four political issues, containing 24,658 posts. We use the same splits as \cite{li2018}. Most previous work reports accuracy. However, given that labels are highly imbalanced, we also report macro F1. 

\paragraph{Learning and Representation:} We model the problem as a collective classification task by predicting disagreement between consecutive posts in a given thread. We represented posts using a BERT encoder. For disagreement, we just represented pairs of posts without additional information. We do 5-fold cross validation and report the average performance. For training, we used AdamW, weight decay of 1e-5, a patience of 3, and 50 restarts for randomized inference. For local models, we used a learning rate of 5e-5 and for structured models we used a learning rate of 2e-6. For structured learning, we initialize the parameters using the local models. Note that we keep fine-tuning BERT during training.

\begin{table}[h]
    \centering
    \resizebox{\columnwidth}{!}{%
        \begin{tabular}{llcc|cc|cc|cc}
        \toprule
             Model & \multicolumn{1}{c}{Infer.}  & \multicolumn{2}{c}{A} & \multicolumn{2}{c}{E} & \multicolumn{2}{c}{GM} & \multicolumn{2}{c}{GC} \\ 
             \midrule

            \multicolumn{2}{c}{}  & Acc & F1
            & Acc & F1
            & Acc & F1
            & Acc & F1 \\
            \midrule
            \multirow{1}{*}{\bf Majority} 
            &           & 56.8 & 28.4 & 65.9 & 33.0 & 66.0 & 33.0 & 67.9 & 34.0 \\  
            \midrule
        \multirow{1}{*}{\bf Local} 
            &          & 66.0 & 64.3 & 65.2 & 54.3 & 70.0 & 61.5 & 68.2 & 54.1 \\

            \midrule
        \multirow{6}{*}{\bf Base}
            & L+I  & 71.0 & 70.4 & 63.3 & 59.2 & 73.6 & 69.4 & 66.8 & 60.2 \\ 
            & ILP         & \textbf{72.4} & \textbf{71.8} & \textbf{64.7} & 60.6 & \textbf{75.1} & 72.6 & 70.5 & 65.8 \\ 
            & AD$^3$/ILP  & \textbf{72.4} & \textbf{71.8} & 63.2 & 59.4 & \textbf{75.1} & 72.6 & 69.7 & 64.3 \\
            & AD$^3$      & \textbf{72.4} & \textbf{71.8} & 64.5 & \textbf{60.7} & \textbf{75.1} & 72.6 & \textbf{71.0} & \textbf{66.0} \\ 
            & Rand-C      & 71.9 & 71.5 & 63.1 & 60.0 & 75.0 & \textbf{73.0} & 65.4 & 60.8 \\
            & Rand        & 71.5 & 71.1 & 61.3 & 58.0 & 74.3 & 72.0 & 64.1 & 60.2 \\  
            \midrule
        
        \multirow{6}{*}{\bf AC}
            & L+I        & 83.6 & 84.6 & 73.3 & 69.7 & 84.8 & 81.9 & 68.2 & 60.9 \\
            & ILP        & 87.5 & \textbf{88.0} & 76.1 & 73.8 & \textbf{91.2} & \textbf{90.3} & 74.2 & 69.9 \\ 
            & AD$^3$/ILP & 86.2 & 85.8 & \textbf{76.7} & \textbf{73.9} & 90.0 & 89.0 & \textbf{74.4} & 70.7 \\  
            & AD$^3$     & 85.0 & 84.8 & 62.7 & 60.3 & 87.4 & 86.3 & 72.8 & 67.9 \\
            & Rand-C     & \textbf{87.8} & 87.6 & \textbf{76.7} & 73.7 & 88.9 & 87.7 & 73.4 & \textbf{71.3} \\ 
            & Rand       & 86.6 & 86.4 & 73.4 & 70.9 & 89.9 & 88.9 & 72.7 & 68.8 \\ 
            \bottomrule
        \end{tabular}
    }
    \caption{\textbf{Post stance} on \fourforums.
    A: Abortion, E: Evolution, GM: Gay Marriage, GC: Gun Control}
    \label{tab:poststance}
\end{table}%
    
\begin{table}
    \centering
    \resizebox{\columnwidth}{!}{%
        \begin{tabular}{llcc|cc|cc|cc}
            \toprule
             Model & \multicolumn{1}{c}{Infer.}  & \multicolumn{2}{c}{A} & \multicolumn{2}{c}{E} & \multicolumn{2}{c}{GM} & \multicolumn{2}{c}{GC} \\ 
             \midrule
           
            \multicolumn{2}{c}{}  & Acc & F1
            & Acc & F1
            & Acc & F1
            & Acc & F1 \\
            \midrule
        \multirow{1}{*}{\bf Majority} 
            &           & 77.8 & 38.9 & 66.4 & 33.2 & 73.7 & 36.9 & 64.3 & 32.2 \\  
            \midrule
        \multirow{1}{*}{\bf Local} 
            &          & 76.0 & 58.1 & 63.4 & 56.0 & 71.3 & 56.9 & 67.0 & 61.4 \\ 
            
            \midrule
        \multirow{6}{*}{\bf Base}
            & L+I        & 70.8          & 60.3          & 62.6          & 58.4          & 63.3          & 58.7          & 61.4          & 58.9 \\ 
            & ILP        & 74.2          & \textbf{62.9} & 63.6          & \textbf{59.6} & \textbf{71.5} & 61.4          & 63.8          & 59.9 \\ 
            & AD$^3$/ILP & 74.2          & \textbf{62.9} & 62.8          & 58.8          & \textbf{71.5} & 61.4          & 64.2          & 61.5 \\ 
            & AD$^3$     & 74.2          & \textbf{62.9} & \textbf{64.3} & 59.5          & \textbf{71.5} & 61.4          & 64.2          & 59.9 \\
            & Rand-C     & \textbf{76.0} & 61.6          & 64.1          & 57.3          & 71.2          & 60.1          & 64.8          & \textbf{61.8} \\
            & Rand       & \textbf{76.0} & 60.7          & 62.7          & 58.5          & 70.4          & \textbf{61.5} & \textbf{65.3} & 59.4 \\
            \midrule
        
        \multirow{6}{*}{\bf AC}
            & L+I         & 83.2 & 78.7 & 72.1 & 70.0 & 71.0 & 68.0 & 68.8 & 67.0 \\ 
            & ILP         & 88.0 & 82.2 & \textbf{76.5} & \textbf{73.6} & \textbf{86.1} & \textbf{81.5} & 75.4 & 73.6 \\ 
            & AD$^3$/ILP  & 86.3 & 80.5 & 74.8 & 72.4 & 83.9 & 79.2 & 72.8 & 71.5 \\ 
            & AD$^3$      & 84.9 & 76.4 & 66.8 & 61.4 & 84.4 & 78.4 & 68.1 & 65.8 \\ 
            & Rand-C      & \textbf{88.2} & \textbf{82.4} & 74.3 & 71.7 & 82.5 & 78.1 & \textbf{78.5} & \textbf{76.3} \\ 
            & Rand        & 87.7 & 81.4 & 75.0 & 71.9 & 84.1 & 78.7 & 75.8 & 74.4 \\ 
            \bottomrule
        \end{tabular}
   }
    \caption{\textbf{Disagreement} on \fourforums. 
    A: Abortion, E: Evolution, GM: Gay Marriage, GC: Gun Control}
    \label{tab:disagreement}
\end{table}

We test two versions of the model: (1) \textbf{Base} includes consistency between node and edge labels, and (2) \textbf{AC} adds author constraints enforcing the same stance for all posts by the same author. 

\paragraph{Structured Learning:} We can also see that the performance of all structured models outperforms learning local models and using inference just at prediction time (L+I), both for post stance (Table~\ref{tab:poststance}) and for disagreement (Table~\ref{tab:disagreement}). 

\paragraph{Randomized vs. ILP/AD3:} In the case of stance prediction, there is a significant trend in the performance of the different inference methods. Learning with exact inference generally outperforms the randomized constrained procedure, and the latter outperforms its non-constrained version. The difference is more pronounced in the case of \textbf{AC} models. However, we find that relative to its simplicity, the randomized procedures obtain highly competitive performance. 

\begin{table}[h]\small
    \centering
    \resizebox{\columnwidth}{!}{%
    \begin{tabular}{lccccc}
        \toprule
        Model & A & E & GM & GC  & Avg\\
        \midrule
        BERT \cite{devlin2018bert} & 66.0 & 65.2 & 70.0 & 68.2 & 67.4\\
        PSL \cite{sridhar:acl15}   & 77.0 & 80.3 & 80.5 & 69.1 & 76.7\\
        Struct. Rep. \cite{li2018}* & 86.5 & \textbf{82.2} & 87.6 & \textbf{83.1} & \textbf{84.9}\\
        \midrule
        Deep AC ILP                & 87.5 & 76.1 & \textbf{91.2} & \textbf{74.2} & \textbf{82.3}\\
        Deep AC Rand-C             & \textbf{87.8} & \textbf{76.7} & 88.9 & 73.4 & 81.7\\
        Deep AC Rand               & 86.6 & 73.4 & 89.9 & 72.7 & 80.7\\
        \bottomrule

    \end{tabular}
    }
    \caption{Previous work on \fourforums (Post Acc)\\ \small{*Note that \protect\cite{li2018} use author profile information in their models, whereas we only look at text}}
    \label{tab:4forums_previous_work}
\end{table}

Table~\ref{tab:4forums_previous_work} compares our models to previous work on this dataset. \citeauthor{sridhar:acl15}~\shortcite{sridhar:acl15} use probabilistic soft logic (PSL) to learn a global assignment for the post labels. They use local classifiers to obtain the input scores to PSL. The main difference between their approach and ours is that we are able to backpropagate the global error back into the classifiers, and we find that it improves performance considerably. Even though we use BERT encoders in our structured procedure, we can see that BERT alone is not able to solve the task. Lastly, we compare to the structured representation learning method of \citeauthor{li2018}~\shortcite{li2018} and find that we are able to improve on abortion and gay marriage only. Note that these two are the issues with more data available (8,000 and 7,000 posts respectively). The main difference with their approach and ours is that they push author profile information into the learned representation. We hypothesize that this is key to obtain good performance for gun control, which contains only 4,000 posts. 




\subsection{Inference Analysis}

In our experiments, randomized inference always outperforms ILP and AD$^3$ in terms of runtime.
Figure~\ref{fig:infer-speedup} shows the speedup factor per epoch against ILP and AD$^3$.
In argument mining, AD$^3$ is faster than ILP, except on our full model, where both perform similarly.
We noticed that ILP consumes a lot of time in initialization and encoding. 
The randomized inference approach is able to predict argumentative structures 9.1x faster than ILP for our base model, and 7.5x faster than AD$^3$ for our full model.
For stance prediction on \fourforums, ILP is considerably faster than AD$^3$, we presume that this is due to the fact that Gurobi is a highly optimized commercial software, and our graphs are small. Randomized inference is 11x faster than ILP on the base model and beats AD$^3$ by a factor of 27 when author constraints are used. 

\begin{figure}[t]
    \resizebox{\columnwidth}{!}{%
        \begin{tikzpicture}
    \begin{axis}[
        width=12cm,
        height=6cm,
        major x tick style=transparent,
        ybar=4pt,
        bar width=10pt,
        ymajorgrids=true,
        ylabel={Inference Speedup Factor},
        symbolic x coords={AM Base, AM Full, 4F Base, 4F AC},
        xtick=data,
        xticklabels={Base ArgMining, Full ArgMining, Base \fourforums, AC\\ \fourforums},
        x tick label style={font=\small, text width=4.5em, align=center},
        scaled y ticks=false,
        enlarge x limits=0.2,
        ymin=0,
        ymax=30,
        point meta=rawy,
        log ticks with fixed point,
        log origin=infty,
        legend cell align=left,
        nodes near coords,
        every node near coord/.append style={font=\scriptsize, color=black},
        legend columns=2,
        legend style={
                at={(0.5,1.04)},
                anchor=south,
                column sep=1ex,
        },
        every axis plot/.append style={fill},
        grid style=dashed,
    ]
        \addplot[index of colormap=2 of RdBu-4]
            coordinates {(AM Base, 9.1) (AM Full, 7.3) (4F Base, 11) (4F AC, 8.7)};
        \addlegendentry{\hspace*{0.2em}Randomized vs. ILP\phantom{$^3$}\hspace*{1em}}

        \addplot[index of colormap=3 of RdBu-4]
             coordinates {(AM Base, 5.3) (AM Full, 7.5) (4F Base, 19) (4F AC, 27)};
        \addlegendentry{\hspace*{0.2em}Randomized vs. AD$^3$}
    \end{axis}
\end{tikzpicture}
    }
    \caption{Comparing inference speedup per epoch}
    \label{fig:infer-speedup}
\end{figure}
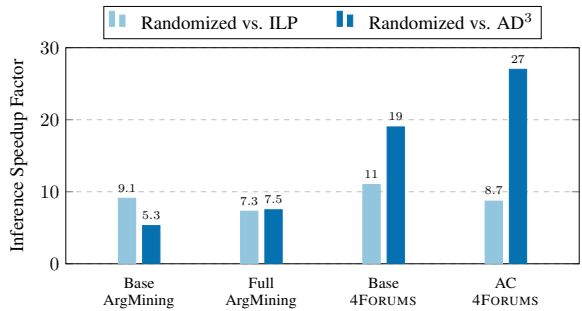

\begin{figure}[t]
    \resizebox{\columnwidth}{!}{%
        \begin{tikzpicture}
    \begin{axis}[
        /pgf/number format/.cd,
        1000 sep={},
        width=15.5cm,
        height=8.8cm,
        label style={font=\Large},
        major x tick style=transparent,
        ybar=5pt,
        bar width=10pt,
        ymajorgrids=true,
        ylabel={Inference in Seconds},
        symbolic x coords={AM Base, AM Full, 4F Base, 4F AC},
        xticklabels={Base ArgMining, Full ArgMining, Base \fourforums, AC \fourforums},
        x tick label style={text width=5em, align=center},
        xtick=data,
        scaled y ticks=false,
        enlarge x limits=0.175,
        ymin=0.4,
        ymax=700,
        ymode=log,
        point meta=rawy,
        log ticks with fixed point,
        log origin=infty,
        legend cell align=left,
        nodes near coords,
        every node near coord/.append style={font=\footnotesize, color=black},
        legend columns=4,
        legend style={
                at={(0.5,1.04)},
                anchor=south,
                font=\Large,
                /tikz/column 2/.style={column sep=3ex}
        },
        every axis plot/.append style={fill},
        grid style=dashed,
    ]
        \addplot[index of colormap=0 of RdBu-4]
            coordinates {(AM Base, 120) (AM Full, 290) (4F Base, 12) (4F AC, 34)};
        \addlegendentry{\hspace*{0.2em}ILP\phantom{$^3$}}

        \addplot[index of colormap=1 of RdBu-4]
            coordinates {(AM Base, 71) (AM Full, 295) (4F Base, 19) (4F AC, 128)};
        \addlegendentry{\hspace*{0.2em}AD$^3$\hspace*{1em}}

        \addplot[index of colormap=2 of RdBu-4]
             coordinates {(AM Base, 12) (AM Full, 33) (4F Base, 0.9) (4F AC, 3.9)};
        \addlegendentry{\hspace*{0.2em}Rand-C\phantom{$^3$}\hspace*{1em}}

        \addplot[index of colormap=3 of RdBu-4]
             coordinates {(AM Base, 17) (AM Full, 28) (4F Base, 0.8) (4F AC, 0.9)};
        \addlegendentry{\hspace*{0.2em}Rand\phantom{$^3$}}
    \end{axis}
\end{tikzpicture}
    }
    \caption{Pure inference runtime in seconds}
    \label{fig:infer-runtime}
\end{figure}
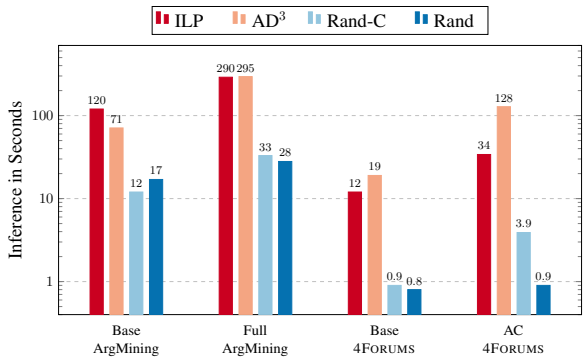

We also measured pure inference time over five training runs and took the average.
Figure~\ref{fig:infer-runtime} shows (in logarithmic scale) plain inference runtime in seconds on the training set for all of our models.
We can observe that randomized inference without domain constraints has almost the same performance as the constrained version.
Again we find that randomized inference considerably outperforms ILP and AD$^3$. 

\begin{figure}[t]
    \centering
    \resizebox{0.55\columnwidth}{!}{
      \begin{tikzpicture} 
    \begin{axis}[%
    hide axis,
    xmin=10,
    xmax=20,
    ymin=0,
    ymax=0.4,
    legend columns=3, 
    legend style={
      font=\Large,
      draw=white!15!black,
      /tikz/column 3/.style={column sep=1pt}
      }
    ]
    \addlegendimage{black, line width=1pt}
    \addlegendentry{\hspace*{0.2em}ILP\hspace*{1em}};

    \addlegendimage{index of colormap=0 of RdBu-4, mark=square, mark size=3pt, line width=1pt}
    \addlegendentry{\hspace*{0.2em}ArgMining\hspace*{1em}};

    \addlegendimage{index of colormap=3 of RdBu-4, mark=triangle, mark size=4pt, line width=1pt}
    \addlegendentry{\hspace*{0.2em}\textsc{4Forums}};
    \end{axis}
\end{tikzpicture}
    }\\[0.2em]

    \begin{minipage}{.5\columnwidth}
        \centering
        \resizebox{\textwidth}{!}{%
            \begin{tikzpicture}
    \begin{axis}[
        xlabel={Number of Restarts},
        ylabel={Avg. F1 (normalized)},
        label style={font=\huge},
        tick label style={font=\huge},
        height=8cm,
        major x tick style=transparent,
        xmin=1, xmax=100,
        ymin=0.83, ymax=1.02,
        xtick={1,20,40,60,80,100},
        ytick={0.85,0.9,0.95,1.0},
        ymajorgrids=true,
        grid style=dashed,
    ]

    \addplot[black, line width=1.5pt] coordinates {(1,1) (100,1)};
    
    \addplot[index of colormap=0 of RdBu-4, smooth, mark=square, mark size=4pt, line width=1.5pt] coordinates {
        (1, 0.906)
        (10, 0.954)
        (20, 0.984)
        (30, 0.990)
        (40, 0.990)
        (50, 0.996)
        (60, 0.997)
        (70, 0.999)
        (80, 0.995)
        (90, 1.000)
        (100, 1.000)
    };
    
    \addplot[index of colormap=3 of RdBu-4, smooth, mark=triangle, mark size=6pt, line width=1.5pt] coordinates {
        (1, 0.841)
        (10, 0.908)
        (20, 0.922)
        (30, 0.935)
        (40, 0.936)
        (50, 0.956)
        (60, 0.940)
        (70, 0.952)
        (80, 0.949)
        (90, 0.957)
        (100, 0.957)
    };
    \end{axis}
\end{tikzpicture}
        }\\
        {\small (a)}
    \end{minipage}\hfill 
    \begin{minipage}{.5\columnwidth}
        \centering
        \resizebox{\textwidth}{!}{%
            \begin{tikzpicture}
    \begin{axis}[
        xlabel={Number of Restarts},
        ylabel={Time (normalized)},
        label style={font=\huge},
        tick label style={font=\huge},
        height=8cm,
        major x tick style=transparent,
        xmin=1, xmax=100,
        ymin=0, 
        xtick={1, 20, 40, 60, 80, 100},
        ytick={0, 0.25, 0.5, 0.75, 1},
        ymajorgrids=true,
        grid style=dashed,
    ]

    \addplot[black, line width=1.5pt] coordinates {(1,1) (100,1)};
    
    \addplot[index of colormap=0 of RdBu-4, smooth, mark=square, mark size=4pt, line width=1.5pt] coordinates {
        (1, 0.09744043825489881)
        (10, 0.16399207232940824)
        (20, 0.24875978404648416)
        (30, 0.33264775325778895)
        (40, 0.4198897937798308)
        (50, 0.5263588303056363)
        (60, 0.5593572314097257)
        (70, 0.6825656479248352)
        (80, 0.7904777400202918)
        (90, 0.8982684702174863)
        (100, 0.9330318489181973)
    };
    
    \addplot[index of colormap=3 of RdBu-4, smooth, mark=triangle, mark size=6pt, line width=1.5pt] coordinates {
        (1, 0.026181258407293583)
        (10, 0.03145105698529412)
        (20, 0.03785011815089806)
        (30, 0.04446590647977942)
        (40, 0.050649642944335944)
        (50, 0.05734845703723384)
        (60, 0.06324225781010646)
        (70, 0.06932791541604436)
        (80, 0.07574118819891239)
        (90, 0.08182366689046225)
        (100, 0.086822135775697)
    };
    \end{axis}
\end{tikzpicture}
        }\\
        {\small (b)}
    \end{minipage}

    \caption{Impact of performance and runtime depending on the number of restarts for randomized inference}
    \label{fig:test-inference}
\end{figure}
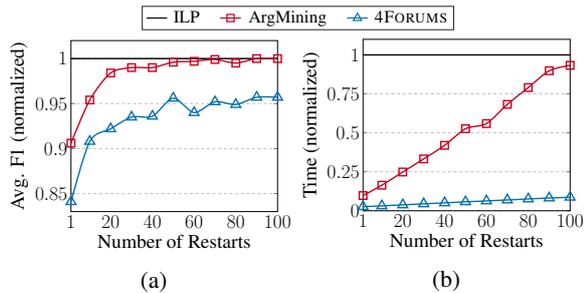

Additionally, we evaluated our model at test time by replacing exact inference with randomized inference, incrementally increasing the number of restarts. 
Figure~\ref{fig:test-inference} shows the performance and runtime of the Rand-C algorithm with respect to exact inference $\left(\text{i.e., }\frac{\text{Rand-C}}{\text{ILP}}\right)$.  
Figure~\extref{fig:test-inference}{a} shows that the global optimum is closely approached after just 20 restarts for the argument mining task, as opposed to stance prediction on \fourforums, where a higher number of restarts is required. This is in line with our reported results in Sections~\ref{sec:argmining_exp} and \ref{sec:4forums_exp}. Figure~\extref{fig:test-inference}{b} shows that randomized inference is about twice as fast than ILP when using 50 restarts for the Argument Mining task, and it starts to approach the time needed for ILP after 100 restarts. On the other hand, the randomized algorithm on \fourforums continues to be an order of magnitude faster even when doing 100 repetitions. Note that as the number of restarts keeps increasing, the randomized procedure will eventually surpass the time needed to perform exact inference.

\section{Summary}\label{sec:summary}

We studied the effectiveness of randomized inference for deep structured prediction and obtained positive results for two challenging discourse-level tasks. We showed that, in practice, we can train complex structured models, using expressive neural architectures, and get competitive results at a lower computational cost. Moreover, we saw that combining expressive representations and inference is a promising direction for modeling discourse-level structures. Future directions include expanding the discussion to other tasks involving more complex structures, as well as exploring shared representations across different sub-tasks. 

\bibliography{eacl2021}
\bibliographystyle{acl_natbib}

\end{document}